\documentclass[10pt,twocolumn,letterpaper]{article}

\usepackage{3dv}
\usepackage{times}
\usepackage{epsfig}
\usepackage{graphicx}
\usepackage{amsmath}
\usepackage{amssymb}

\usepackage{soul}
\usepackage{url}
\usepackage[utf8]{inputenc}
\usepackage[small]{caption}
\usepackage{booktabs}
\usepackage{algorithm}
\usepackage{algorithmic}
\usepackage{amsfonts}
\usepackage{bbm}

\usepackage{multirow}
\usepackage{rotating}

\makeatletter
\def\Hline{%
\noalign{\ifnum0=`}\fi\hrule \@height 1pt \futurelet
\reserved@a\@xhline}
\makeatother

\threedvfinalcopy 

\def\threedvPaperID{239} 

\ifthreedvfinal\fi
\begin{document}

\title{Synthesizing Diverse Lung Nodules Wherever Massively:\\ 3D Multi-Conditional GAN-based CT Image Augmentation for Object Detection}

\author{Changhee Han$^{1,3}$\hspace{1cm}Yoshiro Kitamura$^2$\hspace{1cm}Akira Kudo$^2$\hspace{1cm}Akimichi Ichinose$^2$\hspace{1cm}Leonardo Rundo$^3$\\Yujiro Furukawa$^4$\hspace{1cm}Kazuki Umemoto$^5$\hspace{1cm}Yuanzhong Li$^2$\hspace{1cm}Hideki Nakayama$^1$\\
$^1$The University of Tokyo, Tokyo, Japan\hspace{1cm}$^2$Fujifilm Corporation, Tokyo, Japan\\$^3$University of Cambridge, Cambridge, UK\hspace{1cm}$^4$Jikei University School of Medicine, Tokyo, Japan\\$^5$Juntendo University School of Medicine, Tokyo, Japan\\
{\tt\small han@nlab.ci.i.u-tokyo.ac.jp}
}

\maketitle
\thispagestyle{empty}

\begin{abstract}

Accurate Computer-Assisted Diagnosis, relying on large-scale annotated pathological images, can alleviate the risk of overlooking the diagnosis. Unfortunately, in medical imaging, most available datasets are small/fragmented. To tackle this, as a Data Augmentation (DA) method, 3D conditional Generative Adversarial Networks (GANs) can synthesize desired realistic/diverse 3D images as additional training data. However, no 3D conditional GAN-based DA approach exists for general bounding box-based 3D object detection, while it can locate disease areas with physicians' minimum annotation cost, unlike rigorous 3D segmentation. Moreover, since lesions vary in position/size/attenuation, further GAN-based DA performance requires multiple conditions. Therefore, we propose 3D Multi-Conditional GAN (MCGAN) to generate realistic/diverse $32 \times 32 \times 32$ nodules placed naturally on lung Computed Tomography images to boost sensitivity in 3D object detection. Our MCGAN adopts two discriminators for conditioning: the context discriminator learns to classify real \textit{vs} synthetic nodule/surrounding pairs with noise box-centered surroundings; the nodule discriminator attempts to classify real \textit{vs} synthetic nodules with size/attenuation conditions. The results show that 3D Convolutional Neural Network-based detection can achieve higher sensitivity under any nodule size/attenuation at fixed False Positive rates and overcome the medical data paucity with the MCGAN-generated realistic nodules---even expert physicians fail to distinguish them from the real ones in Visual Turing Test.
\end{abstract}



\section{Introduction}
Accurate Computer-Assisted Diagnosis (CAD), thanks to recent Convolutional Neural Networks (CNNs), can alleviate the risk of overlooking the diagnosis in a clinical environment.
Such great success of CNNs, including diabetic eye disease diagnosis~\cite{gulshan2016development}, primarily derives from large-scale annotated training data to sufficiently cover the real data distribution. However, obtaining and annotating such diverse pathological images are laborious tasks; thus, the massive generation of proper synthetic training images matters for reliable diagnosis. Researchers usually use classical Data Augmentation (DA) techniques, such as geometric/intensity transformations~\cite{ronneberger2015u,milletari2016v}. However, those one-to-one translated images have intrinsically similar appearance and cannot sufficiently cover the real image distribution, causing limited performance improvement; in this regard, thanks to its  good generalization ability, Generative Adversarial Networks (GANs)~\cite{Goodfellow} can generate realistic but completely new samples using many-to-many mappings for further performance improvement; GANs showed excellent DA performance in computer vision, including $21\%$ performance improvement in eye-gaze estimation~\cite{Shrivastava}.

\begin{figure}[t]
  \centering
  \centerline{\includegraphics[width=1\columnwidth]{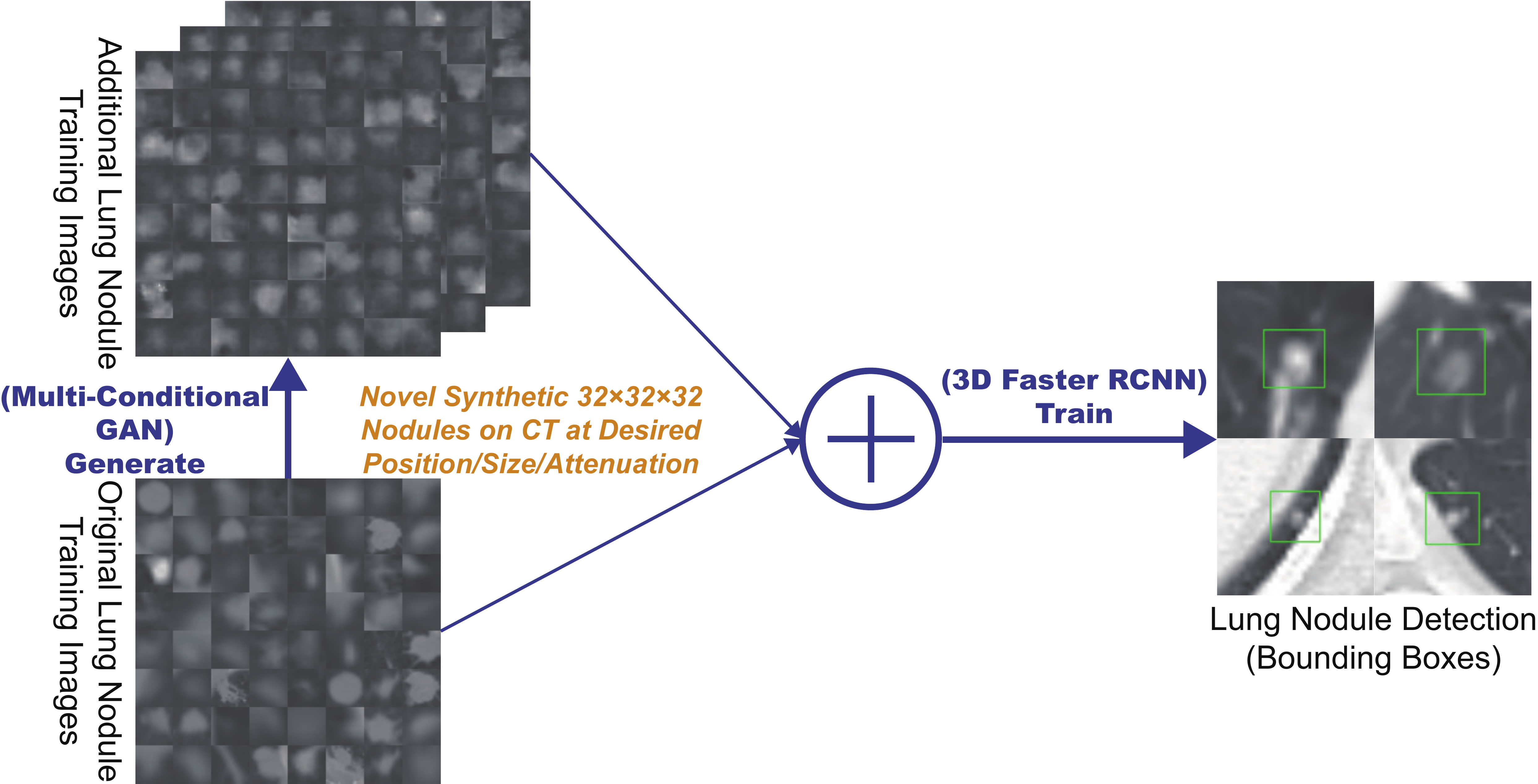}}
\caption{3D MCGAN-based DA for better object detection: Our MCGAN generates realistic and diverse nodules naturally on lung CT scans at desired position/size/attenuation based on bounding boxes, and the CNN-based object detector uses them as additional training data.}
\label{fig1}
\vspace{-6mm}
\end{figure}

This GAN-based DA trend especially applies to medical imaging, where the biggest problem lies in small and fragmented datasets from various scanners. For performance boost in various 2D medical imaging tasks, some researchers used noise-to-image GANs (e.g., random noise samples to diverse pathological images) for classification~\cite{frid2018gan, han2019combining, hanWIRN2018}; others used image-to-image GANs (e.g., a benign image with a pathology-conditioning image to a malignant one) for object detection~\cite{han2019learning} and segmentation~\cite{bailo2019red}. However, although 3D imaging is spreading in radiology (e.g., Computed Tomography (CT) and Magnetic Resonance Imaging), such 3D medical GAN-based DA approaches are limited, and mostly focus on segmentation~\cite{shin2018medical,jin2018ct}---3D medical image generation is more challenging  than 2D one due to expensive computational cost and strong anatomical consistency. Accordingly, no 3D conditional GAN-based DA approach exists for general bounding box-based 3D object detection, while it can locate disease areas with physicians' minimum annotation cost, unlike rigorous 3D segmentation. Moreover, since lesions vary in position/size/attenuation, further GAN-based DA performance requires multiple conditions.

So, how can GAN generate realistic/diverse 3D nodules placed naturally on lung CT with multiple conditions to boost sensitivity in any 3D object detector? For accurate 3D CNN-based nodule detection (Fig.~\ref{fig1}), we propose 3D Multi-Conditional GAN (MCGAN) to generate $32 \times 32 \times 32$ nodules---such nodule detection is clinically valuable for the early diagnosis/treatment of lung cancer, the deadliest cancer~\cite{siegel2019cancer}. Since nodules vary in position/size/attenuation, to improve CNN's robustness, we adopt two discriminators with different loss functions for conditioning: the context discriminator learns to classify real \textit{vs} synthetic nodule/surrounding pairs with noise box-centered surroundings; the nodule discriminator attempts to classify real \textit{vs} synthetic nodules with size/attenuation conditions. We also evaluate the synthetic images' realism \textit{via} Visual Turing Test~\cite{Salimans} by two expert physicians, and visualize the data distribution \textit{via} t-Distributed Stochastic Neighbor Embedding (t-SNE)~\cite{Maaten}. The 3D MCGAN-generated additional training images can achieve higher sensitivity under any nodule size/attenuation at fixed False Positive (FP) rates. Lastly, this study suggests training GANs without $\ell _1$ loss and using proper augmentation ratio (i.e., $1:1$) for better medical GAN-based DA performance.\\


\noindent \textbf{Research Questions.} We mainly address two questions:
\begin{itemize}
\item \textbf{3D Multiple GAN Conditioning:} How can we condition 3D GANs to naturally place objects of random shape, unlike rigorous segmentation, at desired position/size/attenuation based on bounding box masks?
\item \textbf{Synthetic Images for DA:} How can we set the number of real/synthetic training data and GAN loss functions to achieve the best detection performance?
\end{itemize}

\noindent \textbf{Contributions.} Our main contributions are as follows:
\begin{itemize}
\item \textbf{3D Multi-conditional Image Generation:} This first multi-conditional pathological image generation approach shows that 3D MCGAN can generate realistic and diverse nodules placed naturally on lung CT at desired position/size/attenuation, which even expert physicians cannot distinguish from real ones.

\item \textbf{Misdiagnosis Prevention:} This first GAN-based DA method available for any 3D object detector allows to boost sensitivity at fixed FP rates in CAD with limited medical images/annotation.

\item \textbf{Medical GAN-based DA:} This study implies that training GANs without $\ell _1$ loss and using proper augmentation ratio (i.e., $1:1$) may boost CNN-based detection performance with higher sensitivity and less FPs in medical imaging.
\end{itemize}

\section{Generative Adversarial Networks}
GANs~\cite{Goodfellow} have revolutionized image generation~\cite{ledig2017photo} \textit{via} a two-player minimax game.
However, difficult GAN training arises due to its two-player objective function, accompanying artifacts and mode collapse~\cite{DBLP:journals/corr/GulrajaniAADC17} when generating high-resolution images~\cite{Radford}--especially in 3D or conditional image generation; to tackle this, Wu \textit{et al.} proposed 3D GAN~\cite{wu2016learning} to generate realistic/diverse 3D objects \textit{via} a mapping from a low-dimensional probabilistic space; Isola \textit{et al.} proposed \textit{Pix2Pix} GAN~\cite{isola2017image} to produce robust images using paired training samples; Park \textit{et al.} proposed multi-conditional GAN~\cite{park2018mc} to generate $128 \times 128$ images from a base image and texts describing desired position. In this way, GANs can usually synthesize more realistic/diverse images than other common deep generative models, including variational autoencoders~\cite{kingma2013auto} suffering from the injected noise and imperfect reconstruction because of a single objective function~\cite{mescheder2017adversarial}. Accordingly, as a DA method, most computer vision researchers chose GANs for improving classification~\cite{antoniou2017data}, object detection~\cite{ouyang2018pedestrian}, and segmentation~\cite{zhu2018data} to overcome the training data paucity.

\begin{figure*}[t]
  \centering
  \centerline{\includegraphics[width=2\columnwidth]{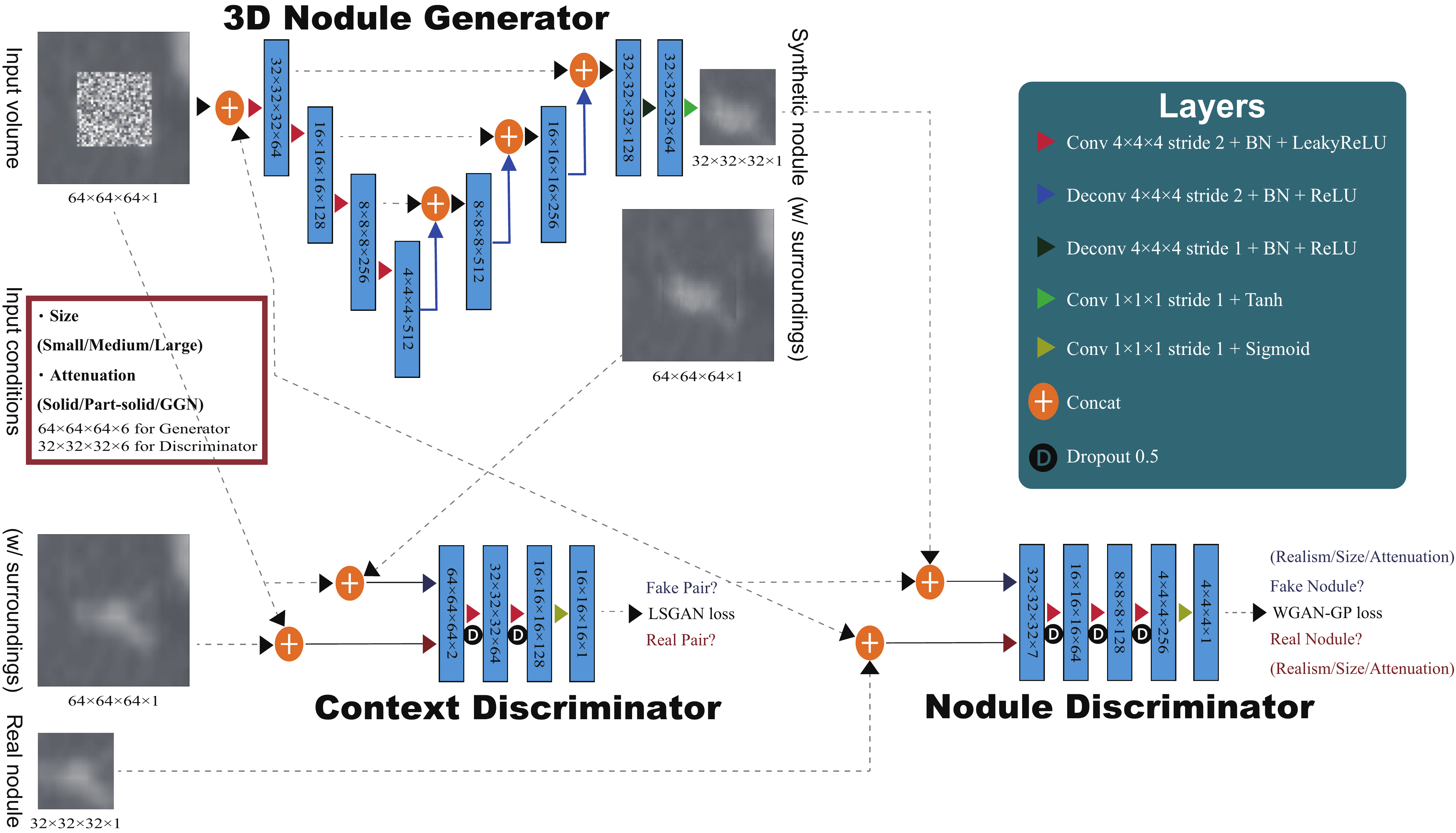}}
\caption{3D MCGAN architecture for realistic and diverse $32 \times 32 \times 32$ lung nodule generation: The context discriminator learns to classify real \textit{vs} synthetic nodule/surrounding pairs while the nodule discriminator learns to classify real \textit{vs} synthetic nodules.}
\label{fig2}
\vspace{-5.7mm}
\end{figure*}

Also in medical imaging, to facilitate object detection and segmentation, researchers usually used conditional GANs to generate medical images at desired positions for DA. Han \textit{et al.} generated $256 \times 256$ brain MR images with tumors at desired positions/sizes for tumor detection~\cite{han2019learning}. As 3D GANs for DA, Jin \textit{et al.}~\cite{jin2018ct} generated $64 \times 64 \times 64$ CT images of both nodules and surrounding tissues---unlike we only generate $32 \times 32 \times 32$ nodules located smoothly on surroundings---for 2D nodule segmentation. Gao \textit{et al.}~\cite{gao2019augmenting} generated $40 \times 40 \times 18$ 3D subvolumes of nodules for subvolume-based 3D nodule detection \textit{via} binary classification; but, the subvolume-based detection accompanies numerous FPs, and unlike our work, most other 3D object detectors cannot use the generated nodules as additional training data since they do not condition nodule positions.

To the best of our knowledge, our work is the first 3D medical GAN-based DA approach using automatic bounding box annotation while 3D bounding boxes require much cheaper annotation cost than rigorous 3D segmentation. Moreover, we, for the first time, generate 3D multi-conditional images using GANs. In terms of annotation cost, generating realistic and diverse $32 \times 32 \times 32$ lung nodules at desired position/size/attenuation using 3D MCGANs---may become a clinical breakthrough .

\section{Methods}

\subsection{3D MCGAN-based Image Generation}
\noindent \textbf{Data Preparation}
This study exploits the Lung Image Database Consortium image collection (LIDC) dataset~\cite{armato2011lung} containing $1,018$ chest CT scans with lung nodules. Since the American College of Radiology recommends lung nodule evaluation using thin-slice CT scans~\cite{setio2017validation}, we only use scans with the slice thickness $\leq 3$ mm and $0.5$ mm $\leq$ in-plane pixel spacing $\leq 0.9$ mm. Then, we interpolate the slice thickness to $1.0$ mm and exclude scans with slice number $> 400$.

To explicitly provide MCGAN with meaningful nodule appearance information and thus boost DA performance, the authors further annotate those nodules by size and attenuation for GAN training with multiple conditions: small (slice thickness $\leq 10$ mm); medium ($10$ mm $\leq$ slice thickness $\leq 20$ mm); large (slice thickness $>$ $20$ mm); solid; part-solid; Ground-Glass Nodule (GGN). Afterwards, the remaining dataset ($745$ scans) is divided into: (\textit{i}) a training set ($632$ scans/$3,727$ nodules); (\textit{ii}) a validation set ($37$ scans/$143$ nodules); (\textit{iii}) a test set ($76$ scans/$265$ nodules); only the training set is used for MCGAN training to be methodologically sound. The training set contains more average nodules since we exclude patients with too many nodules for the validation/test sets; we arrange a clinical environment-like situation, where we could find more healthy patients than highly diseased ones to conduct anomaly detection.

\noindent \textbf{3D MCGAN} is a novel GAN training method for DA, generating realistic but new nodules at desired position/size/attenuation, naturally blending with surrounding tissues (Fig.~\ref{fig2}). We crop/resize various nodules to $32 \times 32 \times 32$ voxels and replace them with noise boxes from a uniform distribution between $[-0.5,0.5]$, while maintaining their $64 \times 64 \times 64$ surroundings as Volumes of Interest (VOIs)---using those noise boxes, instead of boxes filled with the same voxel values, improves the training robustness; then, we concatenate the VOIs with $6$ size/attenuation conditions tiled to $64 \times 64 \times 64$ voxels (e.g., if the size is small, each voxel of the small condition is filled with $1$, while the medium/large condition voxels are filled with $0$ to consider the effect of scaling factor). So, our generator uses the $64 \times 64 \times 64 \times 7$ inputs to generate desired nodules in the noise box regions. The 3D U-Net~\cite{cciccek20163d}-like generator adopts $4$ convolutional layers in encoders and $4$ deconvolutional layers in decoders respectively with skip connections to effectively capture both nodule/context information.

We adopt two \textit{Pix2Pix} GAN~\cite{isola2017image}-like discriminators with different loss functions: the context discriminator learns to classify real \textit{vs} synthetic nodule/surrounding pairs with noise box-centered surroundings using Least Squares loss (LSGAN)~\cite{mao2017least}; the nodule discriminator attempts to classify real \textit{vs} synthetic nodules with size/attenuation conditions using Wasserstein loss with Gradient Penalty (WGAN-GP)~\cite{DBLP:journals/corr/GulrajaniAADC17}. The LSGAN in the context discriminator forces the model to learn surrounding tissue background by reacting more sensitively to every pixel in images than regular GANs. The WGAN-GP in the nodule discriminator allows the model to generate realistic/diverse nodules without focusing too much on details. Empirically, we confirm that such multiple discriminators with the mutually complementary loss functions, along with size/attenuation conditioning, help generate realistic/diverse nodules naturally placed at desired positions on CT scans; similar results are also reported by this work~\cite{ouyang2018pedestrian} for 2D pedestrian detection without label conditioning. We apply dropout to inject randomness and balance the generator/discriminators. Batch normalization is applied to both convolution (using LeakyReLU) and deconvolution (using ReLU).


Most GAN-based DA approaches use reconstruction $\ell _1$ loss~\cite{gao2019augmenting} to generate realistic images, even modifying it for further realism~\cite{jin2018ct}. However, no one has ever validated whether it really helps DA---it assures synthetic images resembling the original ones, sacrificing diversity; thus, to confirm its influence during classifier training, we compare our MCGAN objective without/with it, respectively:
\begin{eqnarray}
	G^*&=&\arg\min_{G}\max_{D1, D2}\mathcal{L}_{\textbf{LSGAN}}(G,D1)\nonumber \\&+&\mathcal{L}_{\textbf{WGAN-GP}}(G,D2),\\
		G^*&=&\arg\min_{G}\max_{D1, D2}\mathcal{L}_{\textbf{LSGAN}}(G,D1)\nonumber \\&+&\mathcal{L}_{\textbf{WGAN-GP}}(G,D2) + 100 \mathcal{L}_{\ell_1}(G).
\end{eqnarray}
We set 100 as a weight for the $\ell _1$ loss, since empirically it works well for reducing visual artifacts introduced by the GAN loss and most GAN works adopt the weight~\cite{isola2017image, ouyang2018pedestrian}.

\noindent \textbf{3D MCGAN Implementation Details}
Training lasts for $6,000,000$ steps with a batch size of $16$ and $2.0 \times 10^{-4}$ learning rate for the Adam optimizer. We use horizontal/vertical flipping as DA and flip real/synthetic labels once in three times for robustness. During testing, we augment nodules with the same size/attenuation conditions by applying a random combination to real nodules of width/height/depth shift up to $10\%$ and zooming up to $10\%$ for better DA. As post-processing, we blend bounding boxes' $3$ nearest surfaces from all the boundaries by averaging the values of $6$ nearest voxels/itself for $5$ iterations. We resample the resulting nodules to their original resolution and map back onto the original CT scans to prepare additional training data.

\subsection{Lung Nodule Detection Using 3D Faster RCNN}
\noindent \textbf{3D Faster RCNN} is a 3D version of Faster RCNN~\cite{ren2015faster} using multi-task loss with a $27$-layer Region Proposal Network of 3D convolutional layers, batch normalization layers, and ReLU layers. To confirm the effect of MCGAN-based DA, we compare the following detection results trained on (\textit{i}) $632$ real images without GAN-based DA, (\textit{ii}), (\textit{iii}), (\textit{iv}) with $1\times$/$2\times$/$3\times$ MCGAN-based DA (i.e., $632$/$1,264$/$1,896$ additional synthetic training images) , (\textit{v}), (\textit{vi}), (\textit{vii}) with $1\times$/$2\times$/$3\times$ MCGAN-based DA trained with $\ell _1$ loss. During training, we shuffle the real/synthetic image order. We evaluate the detection performance as follows: (\textit{i}) Free Receiver Operation Characteristic (FROC) analysis, sensitivity as a function of FPs per scan; (\textit{ii}) Competition Performance
Metric (CPM) score~\cite{Niemeijer}, average sensitivity at seven pre-defined FP rates: 1/8, 1/4, 1/2, 1, 2, 4, and 8 FPs per scan---this quantifies if a CAD system can identify a significant percentage of nodules with both very few FPs and moderate FPs.



\noindent \textbf{3D Faster RCNN Implementation Details}
During training, we use a batch size of $2$ and $1.0 \times 10^{-3}$ learning rate ($1.0 \times 10^{-4}$ after $20,000$ steps) for the SGD optimizer with momentum. The input volume size to the network is set to $160 \times 176 \times 224$ voxels. As classical DA, a random combination of width/height/depth shift up to $15\%$ and zooming up to $15\%$ are also applied to both real/synthetic images to achieve the best performance. For testing, we pick the model with the highest sensitivity on validation between $30,000$-$40,000$ steps under Intersection over Union (IoU) threshold $0.25$/detection threshold $0.5$ to avoid severe FPs.

\subsection{Clinical Validation Using Visual Turing Test}
To quantitatively evaluate the realism of MCGAN-generated images, we supply, in a random order, to two expert physicians a random selection of $50$ real and $50$ synthetic lung nodule images with all of 2D axial/coronal/sagittal views at the center. They take four classification tests in ascending order: Test1, 2: real \textit{vs} MCGAN-generated $32 \times 32 \times 32$ nodules, trained without/with $\ell _1$ loss; Test3, 4: real \textit{vs} MCGAN-generated $64 \times 64 \times 64$ nodules with surroundings without/with $\ell _1$ loss. Such Visual Turing Test~\cite{Salimans} can evaluate the visual quality of GAN-generated medical images in a clinical environment, where physicians' specialty is critical~\cite{han2018gan, chuquicusma2018fool}.

\begin{figure*}[t]
  \centering
  \centerline{\includegraphics[width=2.0\columnwidth]{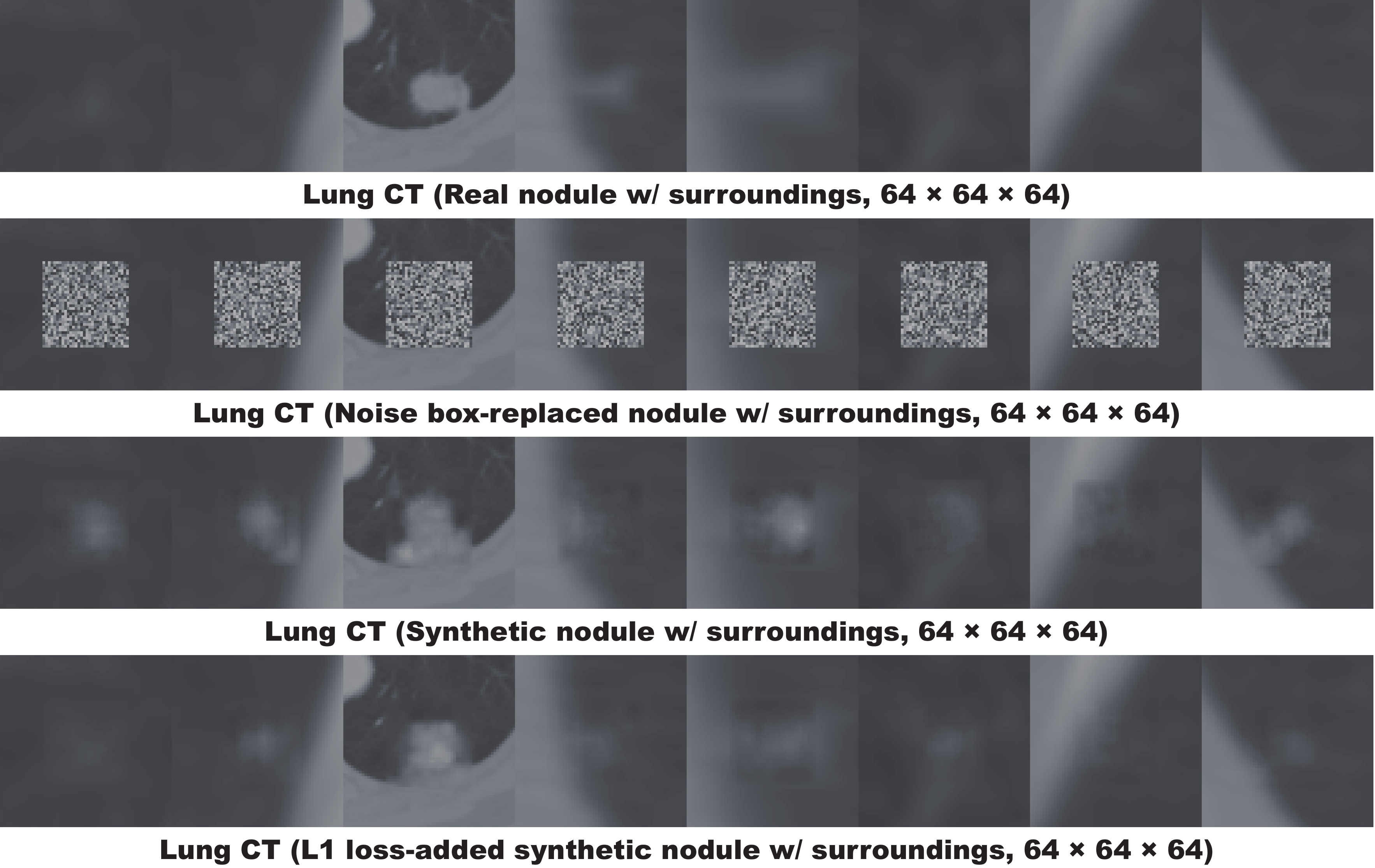}}
\caption{2D axial view of example $64 \times 64 \times 64$ lung nodules with surrounding tissues; 3D MCGANs generate only $32 \times 32 \times 32$ nodules.}
\label{fig3}
\vspace{-1.5mm}
\end{figure*}

\begin{table*}[!t]
\caption{Nodule detection results (CPM) with/without 3D MCGAN-based DA for 3D Faster RCNN under IoU $\geq$ $0.25$. Both results without/with $\ell _1$ loss at different augmentation ratio are compared. CPM is average sensitivity at $1/8, 1/4, 1/2, 1, 2, 4,$ and $8$ FPs per scan.}
\label{tab1}
\centering
\begin{tabular}{lr|rrr|rrr}
\Hline\noalign{\smallskip}
\multicolumn{2}{c}{} & \multicolumn{3}{c}{CPM by Size}& \multicolumn{3}{c}{CPM by Attenuation}\\
& \bfseries  \bfseries CPM & \bfseries Small & \bfseries Medium & \bfseries Large & \bfseries Solid & \bfseries Part-solid & \bfseries GGN \\\noalign{\smallskip}\hline\noalign{\smallskip}
 632 real images & 0.518 & 0.447 & 0.618 & 0.624  & 0.655  & 0.464  & 0.242 \\
 + $1\times$ 3D MCGAN-based DA & \textbf{0.550} & \textbf{0.452} & \textbf{0.683} & \textbf{0.662} & \textbf{0.699} & 0.521 & 0.244\\
 + $2\times$ 3D MCGAN-based DA & 0.527 & 0.447 & 0.674 & 0.429 &	0.655 &	0.407 &	\textbf{0.289} \\
 + $3\times$ 3D MCGAN-based DA & 0.512 & 0.411 & 0.644 & 0.662 &	0.616 &	\textbf{0.579} & 0.277 \\
 + $1\times$ 3D MCGAN-based DA w/ $\ell _1$ & 0.508 & 0.430 & 0.633 & 0.556 & 0.626 & 0.471 & 0.271 \\
 + $2\times$ 3D MCGAN-based DA w/ $\ell _1$ & 0.509 & 0.406 & 0.644 & 0.654 & 0.649 & 0.436 & 0.233 \\
 + $3\times$ 3D MCGAN-based DA w/ $\ell _1$ & 0.479 & 0.389 & 0.594 & 0.617 & 0.596 & 0.507 & 0.226 \\
\noalign{\smallskip}\Hline\noalign{\smallskip}
\end{tabular}
\vspace{-3.2mm}
\end{table*}

\subsection{Visualization Using t-SNE}
To visually analyze the distribution of real/synthetic images, we use t-SNE~\cite{Maaten} on a random selection of $500$ real, $500$ synthetic, and $500$ $\ell _1$ loss-added synthetic nodule images, with a perplexity of $100$ for $1,000$ iterations to get a 2D representation. We normalize the input images to $[0, 1]$. The t-SNE method represents high-dimensional data into a lower-dimensional space by reducing the dimensionality; it uses perplexity to non-linearly balance between the input data's local and global aspects.

\section{Results}
\subsection{Lung Nodules Generated by 3D MCGAN}

We generate realistic nodules in noise box regions at various position/size/attenuation, naturally blending with surrounding tissues including vessels, soft tissues, and thoracic walls (Fig.~\ref{fig3}). Especially, when trained without $\ell _1$ loss, those synthetic nodules look much more different from the original real ones, including slight shading difference.

\begin{figure*}[t]
  \centering
  \centerline{\includegraphics[width=2\columnwidth]{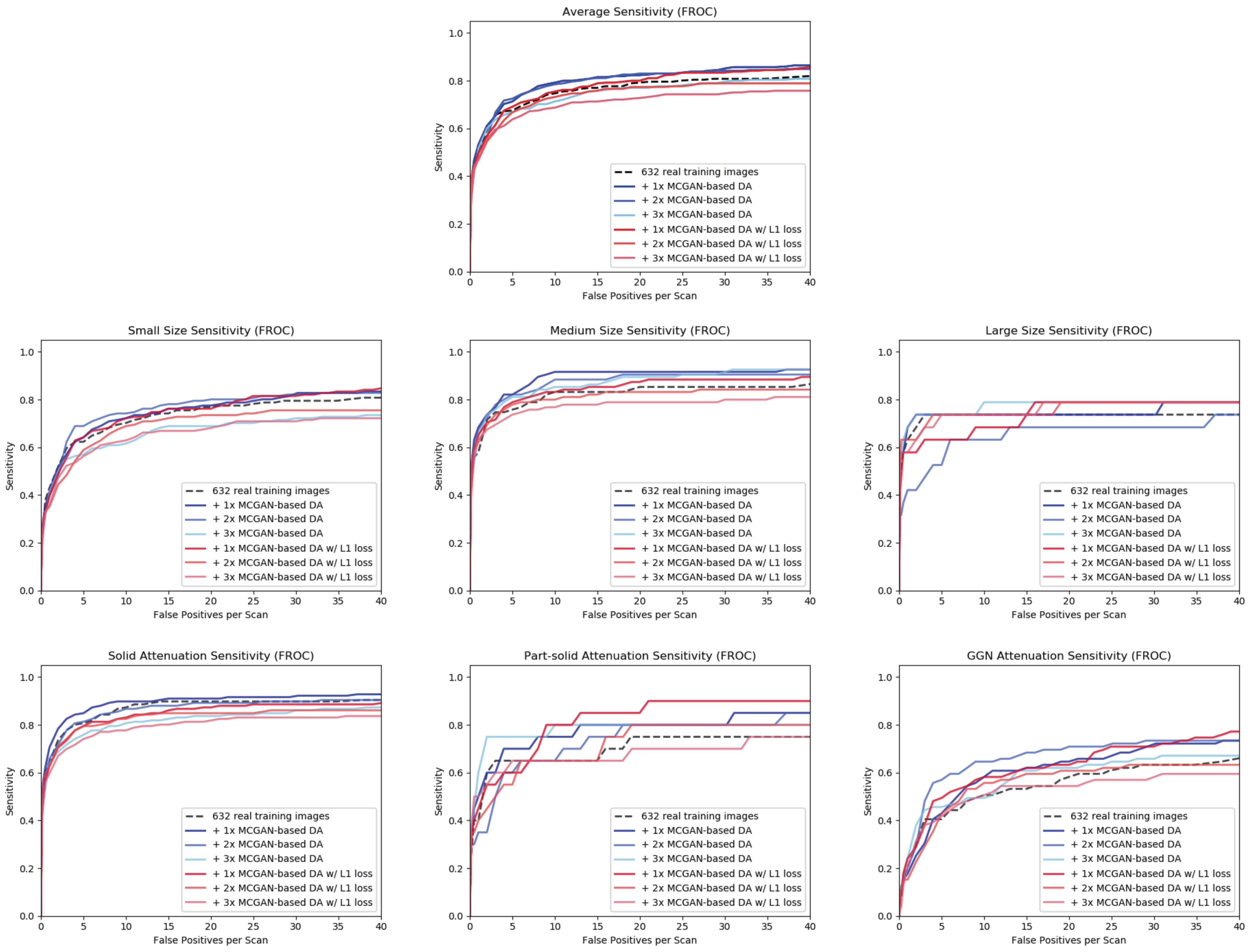}}
\caption{FROC curves of different DA setups by average/size/attenuation.}
\label{fig4}
\end{figure*}

\begin{figure}[t]
  \centering
  \centerline{\includegraphics[width=1\columnwidth]{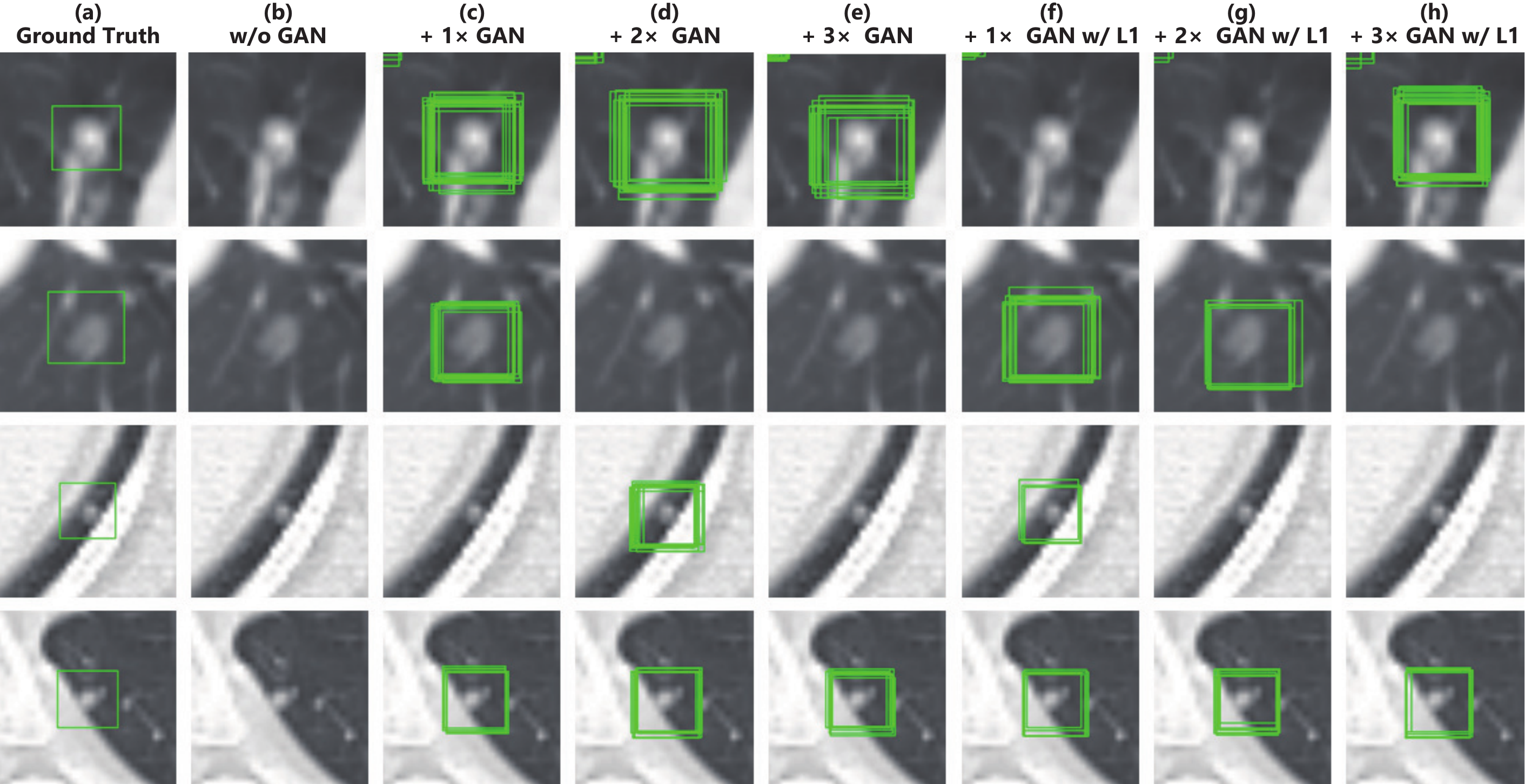}}
\caption{Example detection results with detection threshold $0.5$: (a) ground truth; (b) without GAN-based DA; (c), (d), (e) with $1\times$/$2\times$/$3\times$ 3D MCGAN-based DA; (f), (g), (h) with $1\times$/$2\times$/$3\times$ $\ell _1$ loss-added 3D MCGAN-based DA.}
\label{fig5}
\end{figure}

\subsection{Lung Nodule Detection Results}
Table~\ref{tab1} and Fig.~\ref{fig4} show that it is easier to detect nodules with larger size and lower attenuation due to their clear appearance. 3D MCGAN-based DA with less augmentation ratio consistently increases sensitivity at fixed FP rates---especially, training with $1\times$ MCGAN-based DA without $\ell _1$ loss outperforms training only with real images under any size/attenuation in terms of CPM, achieving average CPM improvement by 0.032. It especially boosts nodule detection performance with larger size and lower attenuation. Fig.~\ref{fig5} visually reveals its ability to alleviate the risk of overlooking the nodule diagnosis with clinically acceptable FPs (i.e., the highly-overlapping bounding boxes around nodules only require a physician's single check by switching on/off transparent alpha-blended annotation on CT scans). Surprisingly, adding more synthetic images tends to decrease sensitivity, probably due to the real/synthetic training image balance. Moreover, further nodule realism introduced by $\ell _1$ loss rather decreases sensitivity as $\ell _1$ loss sacrifices diversity in return for the realism.

\begin{table*}[!t]
\caption{Visual Turing Test results by two physicians for classifying $50$ real \textit{vs} $50$ 3D MCGAN-generated images: Test1, 2: $32 \times 32 \times 32$ lung nodules, trained without/with $\ell _1$ loss; Test3, 4: $64 \times 64 \times 64$ nodules with surrounding tissues, trained without/with $\ell _1$ loss. Accuracy denotes the physicians' successful classification ratio between the real/synthetic images.}
\label{tab2}
\centering
\begin{tabular}{p{1.4em}lrrrrr}
\Hline\noalign{\smallskip}
& \bfseries  & \multicolumn{1}{c}{\bfseries Accuracy} \ \ & \bfseries Real Selected as Real\ \ \  & \bfseries Real as Synt\ \ \  & \bfseries Synt as Real\ \ \  & \bfseries Synt as Synt \\\noalign{\smallskip}\hline\noalign{\smallskip}
\parbox[t]{2mm}{\multirow{2}{*}{\rotatebox[origin=c]{270}{\textbf{\shortstack{\\Test1}}}}} & Physician1 & 43\% \ \ & 19 \ \ & 31 \ \ & 26 \ \ & 24\\
& Physician2 & 43\%\ \ \  & 13\ \ \  & 37\ \ \  & 20\ \ \  & 30\\
\noalign{\smallskip}\hline\noalign{\smallskip}
\parbox[t]{2mm}{\multirow{2}{*}{\rotatebox[origin=c]{270}{\textbf{\shortstack{\\Test2}}}}} & Physician1 & 57\% \ \ & 22 \ \ & 28 \ \ & 15 \ \ & 35\\
& Physician2 & 53\% \ \ & 11 \ \ & 39 \ \ & 8 \ \ & 42\\
\noalign{\smallskip}\hline\noalign{\smallskip}
\parbox[t]{2mm}{\multirow{2}{*}{\rotatebox[origin=c]{270}{\textbf{\shortstack{\\Test3}}}}} & Physician1 & 62\% \ \ & 25 \ \ & 25 \ \ & 13 \ \ & 37\\
& Physician2 & 79\% \ \ & 32 \ \ & 18 \ \ & 3 \ \ & 47\\
\noalign{\smallskip}\hline\noalign{\smallskip}
\parbox[t]{2mm}{\multirow{2}{*}{\rotatebox[origin=c]{270}{\textbf{\shortstack{\\Test4}}}}} & Physician1 & 58\% \ \ & 21 \ \ & 29 \ \ & 13 \ \ & 37\\
& Physician2 & 66\% \ \ & 36 \ \ & 14 \ \ & 20 \ \ & 30\\
\noalign{\smallskip}\Hline\noalign{\smallskip}
\end{tabular}
\end{table*}

\subsection{Visual Turing Test Results}
As Table~\ref{tab2} shows, expert physicians fail to classify real \textit{vs} MCGAN-generated nodules without surrounding tissues---even regarding the synthetic nodules trained without $\ell _1$ loss more realistic than the real ones. Contrarily, they relatively recognize the synthetic nodules with surroundings due to slight shading difference between the nodules/surroundings, especially when trained without the reconstruction $\ell _1$ loss. Considering the synthetic images' realism, CPGGANs might perform as a tool to train medical students and radiology trainees when enough medical images are unavailable, such as abnormalities at rare position/size/attenuation. Such GAN applications are clinically promising~\cite{finlayson2018towards}.

\subsection{T-SNE Results}
\begin{figure}[t]
  \centering
  \centerline{\includegraphics[width=1\columnwidth]{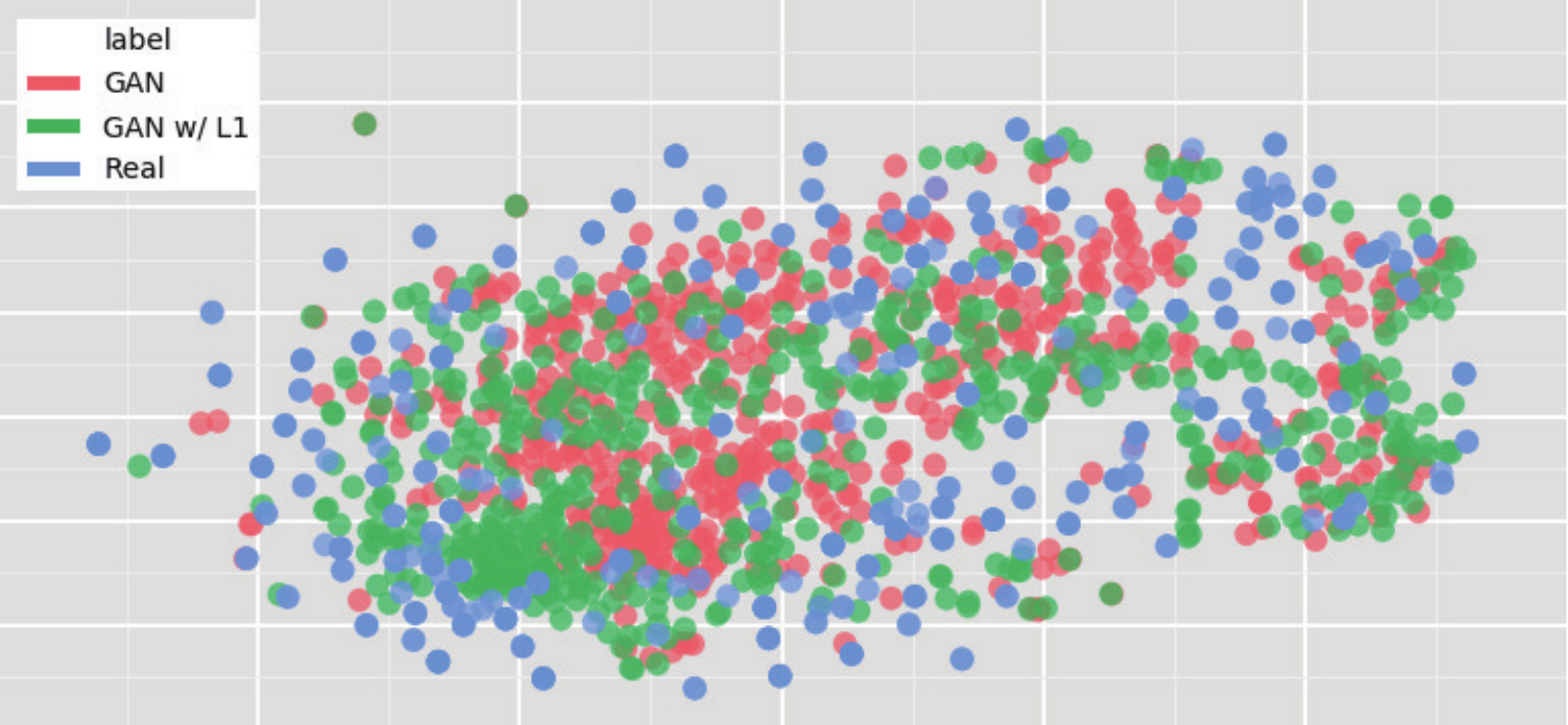}}
\caption{T-SNE plot with 500 $32 \times 32 \times 32$ lung nodule images per each category: (a), (b) 3D MCGAN-generated nodules, trained without/with $\ell _1$ loss; (c) real nodules.}
\label{fig6}
\end{figure}

Implying their  effective  DA  performance, synthetic nodules have a similar distribution to real ones, but concentrated in left inner areas with less real ones especially when trained without $\ell_1$ loss (Fig.~\ref{fig6})--using only GAN loss during training can avoid overwhelming influence from the real image samples, resulting in a moderately similar distribution; thus, those synthetic images can partially fill the real image distribution uncovered by the original dataset.

\section{Conclusion}
Our bounding box-based 3D MCGAN can generate diverse CT-realistic nodules at desired position/size/attenuation, naturally blending with surrounding tissues---those synthetic training data boost sensitivity under any size/attenuation at fixed FP rates in 3D CNN-based nodule detection. This attributes to the MCGAN's good generalization ability coming from multiple discriminators with mutually complementary loss functions, along with informative size/attenuation conditioning; they allow to cover real image distribution unfilled by the original dataset, improving the training robustness.

Surprisingly, we find that adding over-sufficient synthetic images produces worse results due to the real/synthetic image balance; as t-SNE results show, the synthetic images only partially cover the real image distribution, and thus GAN-overwhelming training images rather harm training. Moreover, we notice that GAN training without $\ell _1$ loss obtains better DA performance thanks to increased diversity providing robustness; also expert physicians confirm their sufficient realism without $\ell _1$ loss.

Overall, our 3D MCGAN could help minimize expert physicians' time-consuming annotation tasks and overcome the general medical data paucity, not limited to lung CT nodules. As future work, we will investigate the MCGAN-based DA results without size/attenuation conditioning to confirm their influence on DA performance. Moreover, we will compare our DA results with other non-GAN-based recent DA approaches, such as mixup~\cite{zhang2017mixup} and cutout~\cite{devries2017improved}. For further performance boost, we plan to directly optimize the detection results for MCGANs, instead of realism, similarly to the three-player GAN for classification~\cite{vandenhende2019three}. Lastly, we will investigate how our MCGAN can perform as a physician training tool to display random realistic medical images
with desired abnormalities (i.e., position/size/attenuation conditions) to help train medical students and radiology trainees despite infrastructural and legal constraints~\cite{finlayson2018towards}.

{\small
\bibliographystyle{ieee}
\bibliography{egbib}
}

\end{document}